\documentclass[conference]{IEEEtran}
\IEEEoverridecommandlockouts

\usepackage{cite}
\usepackage{amsmath,amssymb,amsfonts}
\usepackage{algorithmic}
\usepackage{graphicx}
\usepackage{url}
\usepackage{textcomp}
\usepackage{xcolor}
\def\BibTeX{{\rm B\kern-.05em{\sc i\kern-.025em b}\kern-.08em
    T\kern-.1667em\lower.7ex\hbox{E}\kern-.125emX}}

\usepackage[nolist,nohyperlinks]{acronym}
\usepackage{multirow}
\usepackage{tabularx}
\usepackage{subcaption}
\usepackage{svg}

\usepackage{xspace}
\usepackage{subcaption}
\usepackage{adjustbox} 
\usepackage{tikz} 
\usepackage{booktabs}
\usepackage[capitalize]{cleveref}

\crefname{figure}{Fig.}{Fig.}
\Crefname{figure}{Fig.}{Fig.}

\newcommand{\siuru}{SIURU\xspace}

\definecolor{Gray}{gray}{0.9}
\definecolor{Yellow}{rgb}{1.0, 1.0, 0.2}
\definecolor{DarkGreen}{rgb}{0.0, 0.5, 0.0}
\newcommand{\rev}[1]{{\color{black}#1}}
\newcommand{\revv}[1]{{\color{black}#1}}
\newcommand{\revvv}[1]{{\color{black}#1}}

\begin{document}

\title{Binary Anomaly Detection in Streaming IoT Traffic under Concept Drift}



\author{\IEEEauthorblockN{Rodrigo Matos Carnier}
\IEEEauthorblockA{\textit{National Institute of Informatics}\\
Tokyo, Japan\\
rodrigo\_carnier@nii.ac.jp}
\and
\IEEEauthorblockN{Laura Lahesoo}
\IEEEauthorblockA{\textit{Technical University of Munich}\\
Munich, Germany\\
laura.lahesoo@tum.de}
\and
\IEEEauthorblockN{Kensuke Fukuda}
\IEEEauthorblockA{\textit{National Institute of Informatics}\\
Tokyo, Japan\\
kensuke@nii.ac.jp}
}

\maketitle

\begin{acronym}
	\itemsep-.25\baselineskip
    \acro{AD}{anomaly detection}
    \acro{ML}{machine learning}
    \acro{IoT}{Internet of Things}
    \acro{IDS}{intrusion detection system}
    \acro{SIURU}{Scalable IoT Usage Research Utility}
    \acro{XAI}{eXplainable Artificial Intelligence}
    \acro{RF}{Random Forest}
    \acro{ARF}{Adaptive Random Forest}
    \acro{Hoeffding}{Hoeffding Adaptive Tree}
    \acro{DT}{Decision Tree}
    \acro{VFDT}{Very Fast Decision Tree}
    \acro{NB}{Naive Bayes}
    \acro{AE}{Autoencoder}
    \acro{CNN}{Convolutional Neural Network}
    \acro{IF}{Isolation forest}
    \acro{AUC}{Area Under the Curve}
\end{acronym}


\begin{abstract}\label{sec:abstract}
With the growing volume of \ac{IoT} network traffic, \ac{ML}-based anomaly detection is more relevant than ever. Traditional batch learning models face challenges such as high maintenance and poor adaptability to rapid anomaly changes, known as concept drift. In contrast, streaming learning integrates online and incremental learning, enabling seamless updates and concept drift detection to improve robustness.
This study investigates anomaly detection in streaming IoT traffic \revv{as binary classification,} comparing batch and streaming learning approaches while assessing the limitations of current IoT traffic datasets. We simulated heterogeneous network data streams by carefully mixing existing datasets and streaming the samples one by one. Our results highlight the failure of batch models to handle concept drift, but also reveal persisting limitations of current datasets to expose model limitations due to low traffic heterogeneity.
We also investigated the competitiveness of tree-based ML algorithms, well-known in batch anomaly detection, and compared it to non-tree-based ones, confirming the advantages of the former. Adaptive Random Forest achieved F1-score of \rev{0.990 ± 0.006 }at one-third the computational cost of its batch counterpart. Hoeffding Adaptive Tree reached F1-score of \rev{0.910 ± 0.007,} reducing computational cost by \rev{four} times, making it a viable choice for online applications despite a slight trade-off in stability.

\end{abstract}



\section{Introduction}
\label{sec:introduction}

For decades, network traffic anomaly detection has relied on ML techniques \cite{rafique2024,chandola2009}. Early intrusion detection systems (IDS) used heuristic or statistical methods, but after enough computational power became available, security systems adopted ML and evolved into more generalized anomaly detectors, identifying both malicious attacks and system failures by their traffic features.

The age of IoT has made traffic anomaly detection more challenging by increasing the variety of devices, services, anomaly types, and their rate of change. As more IoT devices and services are deployed, new attack surfaces and failure points appear, making it harder to train general ML-based anomaly detection systems \cite{hamza2021, vignau2019}. Datasets quickly become outdated due to growing traffic diversity and the fast emergence of new threats, complicating model maintenance and the creation of updated benchmarks. Studies also highlight the deficiencies of public IoT network traffic datasets \cite{dekeersmaeker2023}.

In batch learning, the traditional ML approach, detection is improved by preprocessing data and extracting key features during offline training. While effective in some contexts, this method struggles with IoT traffic because it requires retraining models with accumulated data to address both old and new threats. As datasets grow, model maintenance slows down. Infrequent updates also reduce model effectiveness against evolving attacks.

Streaming learning is a promising alternative that allows incremental model updates without processing entire datasets, simplifying model and dataset maintenance. It also possesses stronger robustness to concept drift---changes in data distribution that impact ML predictions. Concept drift can occur when certain attacks become obsolete or more advanced ones emerge, or when sudden user behavior shifts affect normal traffic patterns. Because streaming models perform online incremental learning, they can quickly adjust their structure and parameters through concept drift detectors, staying effective even as data distributions evolve.

\revv{However, it is currently challenging to evaluate the real limitations of batch learning and the advantages of streaming learning in the context of IoT traffic, because current traffic datasets lack heterogeneity \cite{poisson2024gothx} -- e.g. parametrically rich and diverse traffic from recent/realistic attacks and benign data representative of diversity of IoT devices and services. With this in mind, this paper contributes with the following:}


\begin{enumerate}
\item A method for simulation of heterogeneous data streams containing multiple concept drifts, using data from different datasets (\cref{subsec:simulation}).
\item An analysis of the robustness of batch and streaming learning methods against concept drift in IoT traffic anomaly detection, \rev{performed as binary classification} (\cref{subsec:robustness}).
\item A comparison of performance and computational cost between tree-based and non-tree-based ML algorithms for IoT traffic anomaly detection (\cref{subsec:algorithm_comparison}).
\end{enumerate}


Our results confirm that batch learning struggles with concept drift while streaming learning shows fast adapting performance. Moreover, we show that current IoT traffic datasets are too homogeneous when used alone, allowing batch ML models to generalize their training data to new anomalies even after concept drift. After further treatment of data to increase heterogeneity, this artificial generalization disappeared, decreasing the performance of batch models significantly after the occurrence of concept drifts. We also confirm the higher competitiveness of tree-based algorithms and suggest the usage of different algorithms for different requirements.



\section{Related Works}
\label{sec:relatedwork}


Studies on anomaly detection for network traffic have shown the advantages of incremental learning over batch learning. However, the implementation of streaming anomaly detection with concept drift adaptation — as opposed to mere online solutions adapted from offline anomaly detection — is still underway \cite{shahraki2022}. IoT data presents additional challenges for streaming learning due to the complexity of anomalous traffic, which arises from data temporal relationships, diverse IoT applications, and faulty hardware \cite{cook2020}.

KITSUNE \cite{KitsuneMirsky2018} is a well-known online semi-supervised Intrusion Detection System (IDS) based on autoencoders, designed for low-power devices like the Raspberry Pi. It has shown an F1-score between 0.8 and 0.99 across nine datasets. However, its dampened incremental statistics method for feature extraction may struggle with more complex concept drift, making it an online incremental learner rather than a fully streaming solution. Raeiszadeh et.al \cite{raeiszadeh2024} proposed a LSTM model for time-series IoT traffic anomaly detection with GA-based hyperparameters optimization. The model performs offline training and real-time testing with concept drift detection, achieving an AUC of 89.71\%. Despite its drift detection capability, the offline training phase limits the benefits of streaming learning. Hussein et.al \cite{hussein2024} proposed a LSTM autoencoder solution for anomaly detection in IoT traffic, achieving an F1 score of 0.847 on the SKAB dataset. However, there was not an evaluation of robustness to concept drift, or considerations on computational cost. 

Considering the lack of studies and comparisons between streaming ML algorithms for anomaly detection, this work will focus on its merits and trade-offs.










\section{Evaluation Setup}
\label{sec:setup}

\revvv{The setup of our study was based on three elements: the selection of datasets containing IoT network traffic, the development of an environment capable of batch and streaming anomaly detection -- including selection of traffic features, evaluation of different ML algorithms, optimization of hyperparameters and application of concept drift detectors -- and more importantly, the preparation of data that contain time-dependent properties related to concept drift and display a more realistic heterogeneous data.
}

\subsection{Datasets}
\label{subsec:datasets}

Due to the highly variable nature of IoT data, we mixed three datasets with MQTT traffic, a publish-subscribe network protocol designed for communication between devices, widely used in IoT applications. The mixing process is described in \cref{subsec:dataprep}.

\textbf{MQTTSet} \cite{MQTTsetPaper2020}: A widely used IoT dataset, generated through a testbed with MQTT service. It offers 5 different types of malicious attacks, split into different pcap files, and legitimate traffic from 10 different IoT sensors all mixed in the same file. 

\textbf{Edge-IIoTset} \cite{EdgesetPaper2022}: A rich cybersecurity dataset designed for IoT applications, generated through a testbed with MQTT and other services like cloud computing, blockchain, and edge computing. It offers 15 different types of malicious attacks and 10 types of IoT sensors, split into different pcap files. It is the most rich public IoT dataset found to date. 

\textbf{MQTT-IoT-IDS2020} \cite{IoTsetPaper2021}: Another MQTT-focused dataset that simulates an IoT network of sensors, cameras, and attackers. It offers 4 types of malicious attacks and a simple file of benign traffic collected from MQTT sensors. 

\subsection{Anomaly detector}
\label{subsec:anomalydetector}

\begin{figure}[t]
\centering
\includegraphics[scale=0.39]{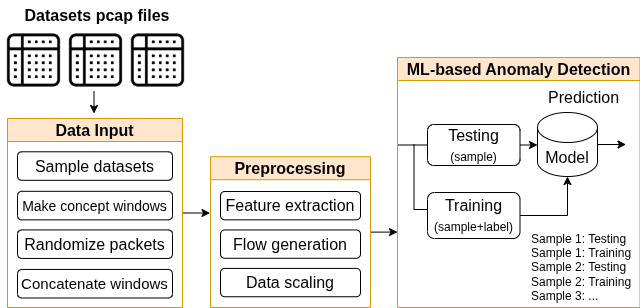}
\caption[SIURU architecture]{Overview of simulation setup and anomaly detector.}
\label{fig:siuru-arch}
\end{figure}

To evaluate streaming anomaly detection, we expanded our in-house anomaly detector, \ac{SIURU}~\cite{Lahesoo2023} and included streaming ML library \textit{River}. 

SIURU's three main components for streaming anomaly detection are shown in \cref{fig:siuru-arch}. ``Data Input'' generates an artificial data stream with four phases of mixed malicious and benign traffic (more details in~\cref{subsec:datasets}). ``Preprocessing'' extracts \revv{a set of 25 features from raw packets in time windows of 1ms. The features include packet-based, host-based and flow-based ones (full list in the previous work~\cite{Lahesoo2023})}. Finally, ``ML-based Anomaly Detection'' detects anomalies sample by sample and incrementally trains the current ML model with the latest tested sample (except for the batch Random Forest method, which is trained offline and not updated). Below is a brief description of the evaluated methods:

\textbf{Batch Random Forest (RF):} An ensemble learning method that constructs multiple decision trees using bootstrap sampling and majority voting. It is a very robust, optimized ML algorithm for batch learning, 
but has the trade-off of being much slower than other ML algorithms.

\textbf{Adaptive Random Forest (ARF):} A streaming version of RF that adapts to concept drifts by dynamically replacing underperforming trees. ARF possesses a high degree of granularity for adaptive anomaly detection, arguably making it the best adaptive method, but its computational cost also lays on the higher side.

\textbf{Hoeffding Adaptive Tree (Hoeff):} Also called Very Fast Decision Tree (VFDT). It efficiently builds and adapts trees for data streams using the Hoeffding bound, which quantifies how much the empirical mean of random variables is likely to deviate from the true mean. This streaming ML algorithm is useful where decisions must be made with limited data.

\textbf{Gaussian Naive Bayes (NB):} A probabilistic classifier based on Bayes’ theorem. It assumes that features follow a Gaussian distribution and are conditionally independent. It is computationally efficient, simple and interpretable, but it loses performance if feature independence is not realistic.

Hyper-parameters for all algorithms were optimized using AdaBound, an adaptive optimization algorithm based on Adam and SGD with momentum, and a multi-armed bandit (MAB) algorithm.

Streaming ML depends on concept drift detectors, which check changes in data distribution and prompt the algorithm to readjust model parameters using only recent data. We used the algorithm ADWIN (ADaptive WINdowing), which uses two sliding windows of data to detect changes in data distribution and adjusts the window size dynamically.

\subsection{Data preparation}
\label{subsec:dataprep}

To compare batch and streaming ML under concept drift, it is necessary to prepare data in specific ways. First, incremental training and testing of the entire sequence of data should simulate a time-series process, as the order of incoming samples significantly affects detection performance---especially since concept drift detectors induce streaming ML algorithms to forget old data. Second, different anomaly phases must be carefully structured to highlight the impact of shifts in data distribution. Third, batch models should test samples in incremental fashion, restricting preprocessing techniques to data scaling, to ensure a fair comparison.

To satisfy these requirements, \revv{we developed realistic simulations of IoT streaming traffic, controlling at which point of the simulation the concept drift happens, and which type of drift. Every simulation possesses four phases of traffic. Each phase contains a mix of benign traffic and a single type of malicious traffic (binary classification per phase). The packets of benign and malicious traffic inside each phase of attack were mixed randomly, without seed. Since the ML anomaly detector is not pretrained to detect the second, third and fourth anomalies, a change in the learned distribution of data occurs everytime a new phase starts.






\begin{table}[t]
    \caption{Structure of the six simulations of streaming IoT traffic, each with three concept drifts.}
    \label{tab:simulations}
    \centering
    \begin{tabular}{cccc}
        \toprule
        \textbf{Simulation} & \textbf{Dataset} & \multicolumn{2}{c}{\textbf{Sequence of attack phases}}\\
        \midrule
        1 & 1 & \multicolumn{2}{c}{E1 $\rightarrow$ M1 $\rightarrow$ E2 $\rightarrow$ M2}\\
        2 & 1 & \multicolumn{2}{c}{M2 $\rightarrow$ E2 $\rightarrow$ M1 $\rightarrow$ E1}\\
        3 & 2 & \multicolumn{2}{c}{M1 $\rightarrow$ I1 $\rightarrow$ M2 $\rightarrow$ I2}\\
        4 & 2 & \multicolumn{2}{c}{I2 $\rightarrow$ M2 $\rightarrow$ I1 $\rightarrow$ M1}\\
        5 & 3 & \multicolumn{2}{c}{E1 $\rightarrow$ I1 $\rightarrow$ E2 $\rightarrow$ I2}\\
        6 & 3 & \multicolumn{2}{c}{I2 $\rightarrow$ E2 $\rightarrow$ I1 $\rightarrow$ E1}\\
        \bottomrule
    \end{tabular}
\end{table}
\begin{table}[t]
    \caption{Properties of the three mixed datasets used in the simulations.}
    \label{tab:datasets}
    \centering
    \begin{tabular}{cccc}
        \toprule
        \multicolumn{4}{c}{\textbf{Mixed dataset 1: Edge + MQTT}}\\
        \multirow{2}{*}{\textbf{Phase}} & \multirow{2}{*}{\textbf{Class}} & \multirow{2}{*}{\textbf{Type}} & \rev{\textbf{Traffic samples}}\\
        & & & \rev{\textbf{(Benign + Malicious)}} \\
        \midrule
        E1 & Malware & Password bruteforce & \rev{2369 + 3072} \\
        M1 & DoS & Malaria DoS &\rev{ 2637 + 2610} \\
        E2 & DoS & HTTT Flood & \rev{2263 + 2021} \\
        M2 & Malware & MQTT bruteforce & \rev{2639 + 2011} \\
        \toprule
        \multicolumn{4}{c}{\textbf{Mixed dataset 2: MQTT + IoT}} \\
        \multirow{2}{*}{\textbf{Phase}} & \multirow{2}{*}{\textbf{Class}} & \multirow{2}{*}{\textbf{Type}} & \rev{\textbf{Traffic samples}}\\
        & & & \rev{\textbf{(Benign + Malicious)}} \\
        \midrule
        M1 & Malware & MQTT bruteforce & \rev{2639 + 2011 }\\
        I1 & Info Gather & Aggressive Scan &\rev{ 1198 + 654} \\
        M2 & DoS & Malaria DoS & \rev{2637 + 2610} \\
        I2 & Malware & MQTT bruteforce & \rev{1021 + 346} \\
        \toprule
        \multicolumn{4}{c}{\textbf{Mixed dataset 3: Edge + IoT}}\\
        \multirow{2}{*}{\textbf{Phase}} & \multirow{2}{*}{\textbf{Class}} & \multirow{2}{*}{\textbf{Type}} & \rev{\textbf{Traffic samples}}\\
        & & & \rev{\textbf{(Benign + Malicious)}} \\
        \midrule
        E1 & Malware & Password bruteforce & \rev{2369 + 3072} \\ 
        I1 & Info Gather & Aggressive Scan & \rev{1198 + 654} \\ 
        E2 & DoS & HTTT Flood & \rev{2263 + 2021} \\ 
        I2 & Malware & MQTT bruteforce & \rev{1021 + 346}\\ 
        \bottomrule
    \end{tabular}
\end{table}

\subsection{Simulation of Streaming Anomaly Detection}
\label{subsec:simulation}

\begin{figure*}[ht]
    \centering
    \captionsetup[subfigure]{font=footnotesize} 

    \begin{minipage}[b]{0.39\linewidth}
        \centering
        \begin{subfigure}[b]{0.48\linewidth}
            \includegraphics[width=\linewidth]{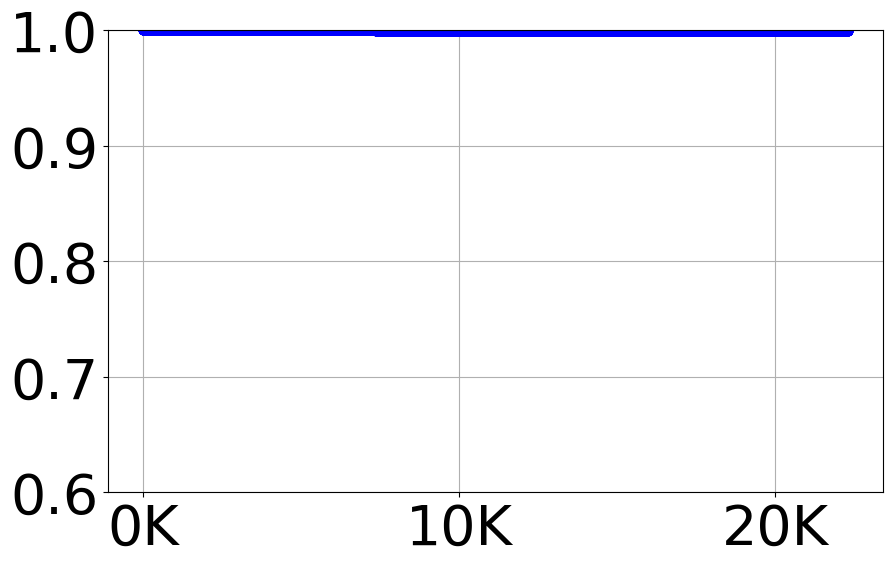}
            \caption{\rev{RF $\rightarrow$ Edge dataset\\(single dataset).}}
            \label{fig:RF-edge}
        \end{subfigure}
        \hfill
        \begin{subfigure}[b]{0.48\linewidth}
            \includegraphics[width=\linewidth]{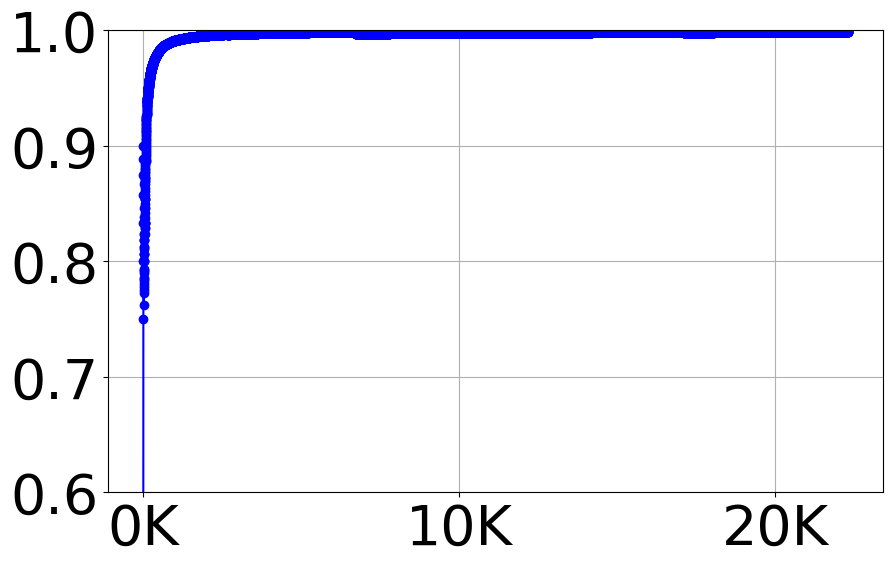}
            \caption{\rev{ARF $\rightarrow$ Edge dataset\\(single dataset).}}
            \label{fig:ARF-edge}
        \end{subfigure}
        
        \vspace{0.5cm} 
        \begin{subfigure}[b]{0.48\linewidth}
            \includegraphics[width=\linewidth]{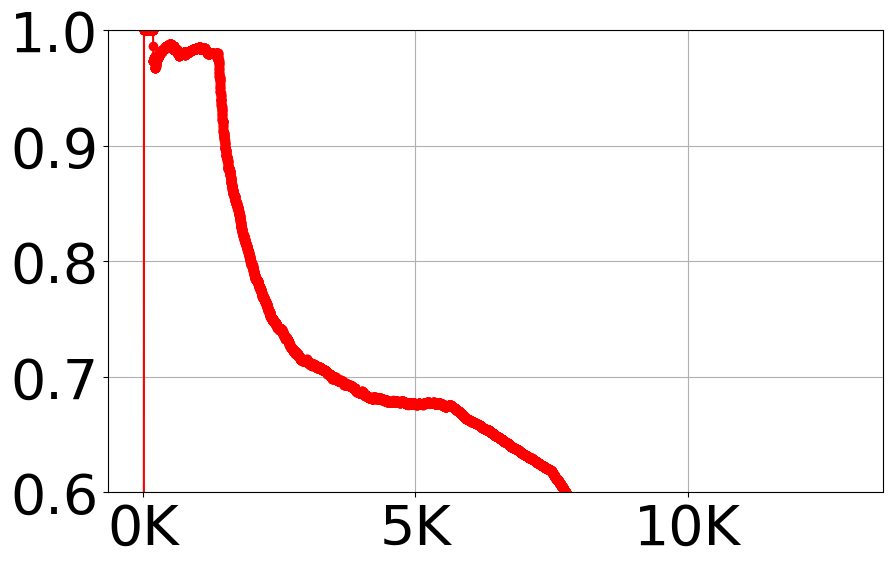}
            \caption{\rev{RF $\rightarrow$ Simulation 6 (mixed dataset).}}
            \label{fig:RF-edgeMQTT}
        \end{subfigure}
        \hfill
        \begin{subfigure}[b]{0.48\linewidth}
            \includegraphics[width=\linewidth]{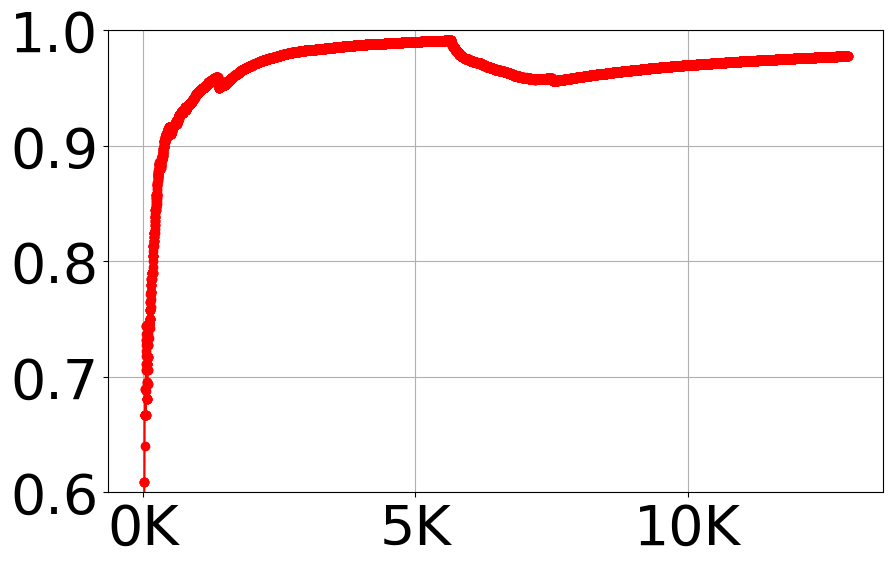}
            \caption{\rev{ARF $\rightarrow$ Simulation 6 (mixed dataset).}}
            \label{fig:ARF-edgeMQTT}
        \end{subfigure}
        
        \caption{Comparison between batch and streaming ML. \rev{Top: single dataset. Bottom: mixed dataset}. X-axis: samples. Y-axis: cumulative \rev{F1-score}.}
        \label{fig:comparison-batch-stream}
    \end{minipage}
    \hfill
    \begin{minipage}[b]{0.59\linewidth}
        \centering
        \begin{subfigure}[b]{0.32\linewidth}
        \includegraphics[width=\linewidth]{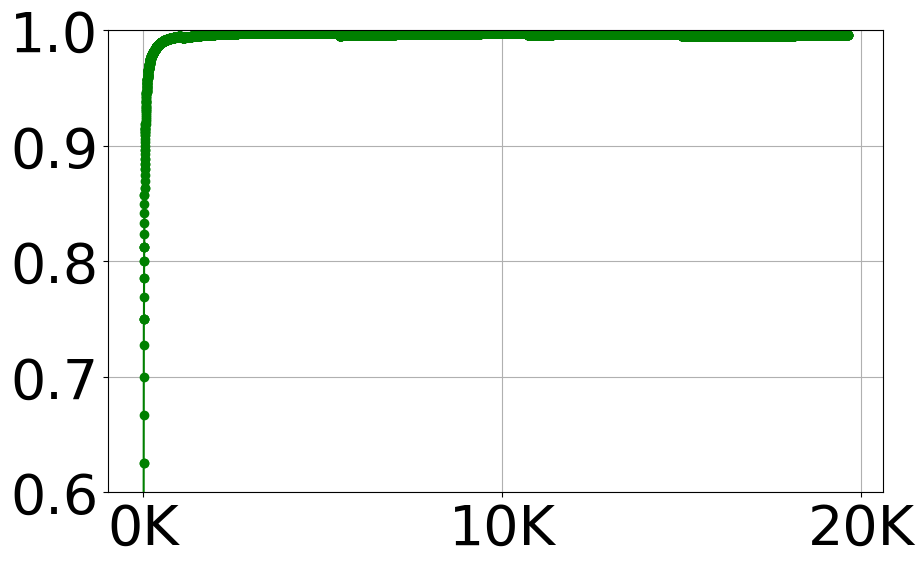}
        \caption{\rev{Best ARF $\rightarrow$ Simulation 1 (mixed dataset).}}
        \label{fig:arf-best}
        \end{subfigure}
        \hfill
        \begin{subfigure}[b]{0.32\linewidth}
        \includegraphics[width=\linewidth]{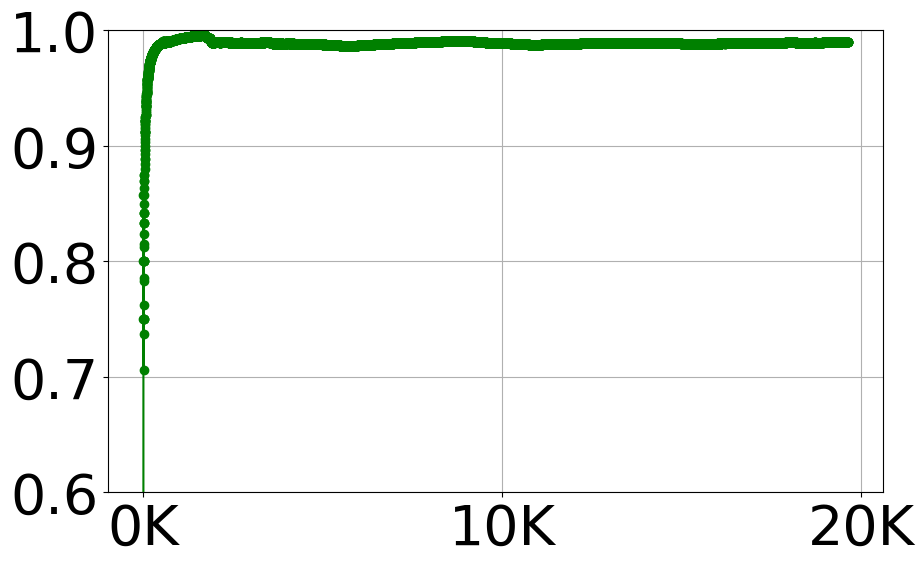}
        \caption{\rev{Best Hoeff $\rightarrow$ Simulation 1 (mixed dataset).}}
        \label{fig:hoeff-best}
        \end{subfigure}
        \hfill
        \begin{subfigure}[b]{0.32\linewidth}
        \includegraphics[width=\linewidth]{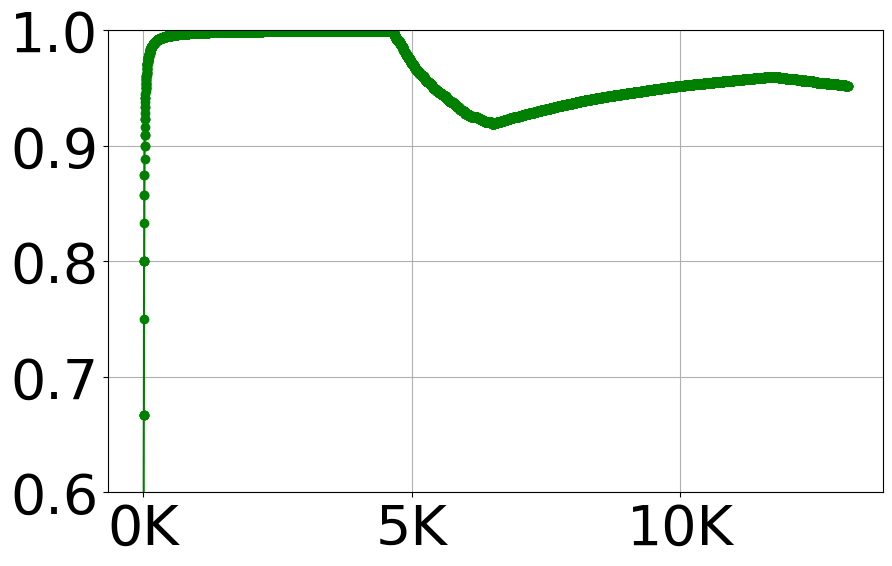}
        \caption{\rev{Best NB $\rightarrow$ Simulation 3 (mixed dataset).}}
        \label{fig:nb-best}
        \end{subfigure}
        
        \vspace{0.5cm} 
        \begin{subfigure}[b]{0.32\linewidth}
        \includegraphics[width=\linewidth]{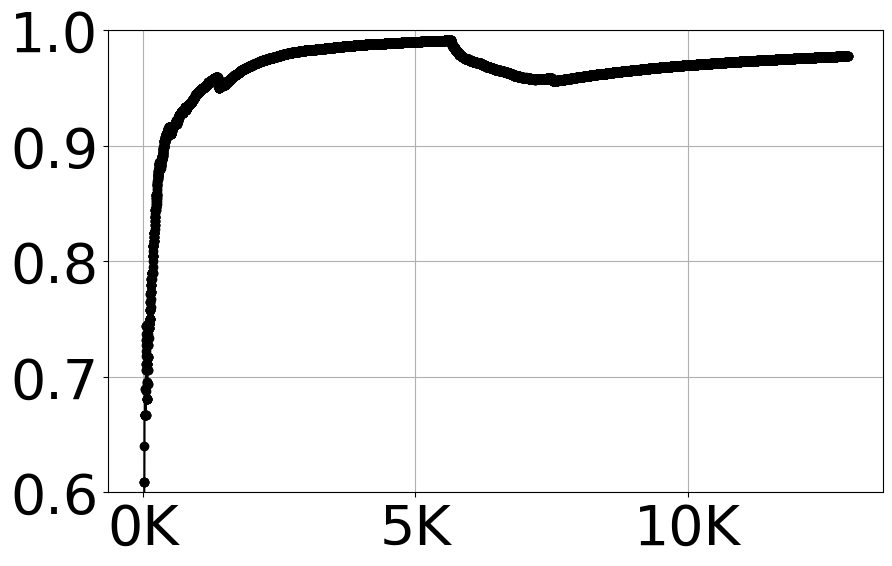}
        \caption{\rev{Worst ARF $\rightarrow$ Simulation 6 (mixed dataset).}}
        \label{fig:arf-worst}
        \end{subfigure}
        \hfill
        \begin{subfigure}[b]{0.32\linewidth}
        \includegraphics[width=\linewidth]{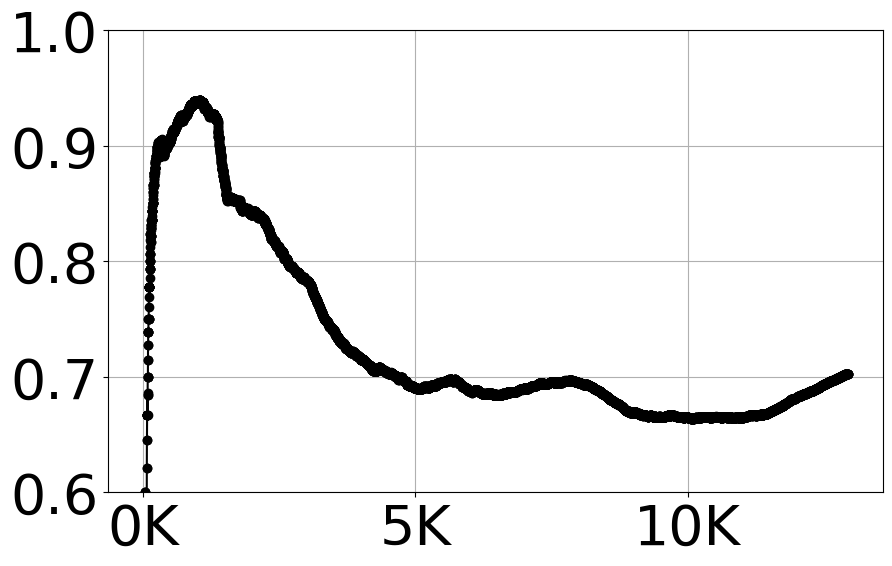}
        \caption{\rev{Worst Hoeff $\rightarrow$ Simulation 6 (mixed dataset).}}
        \label{fig:hoeff-worst}
        \end{subfigure}
        \hfill
        \begin{subfigure}[b]{0.32\linewidth}
        \includegraphics[width=\linewidth]{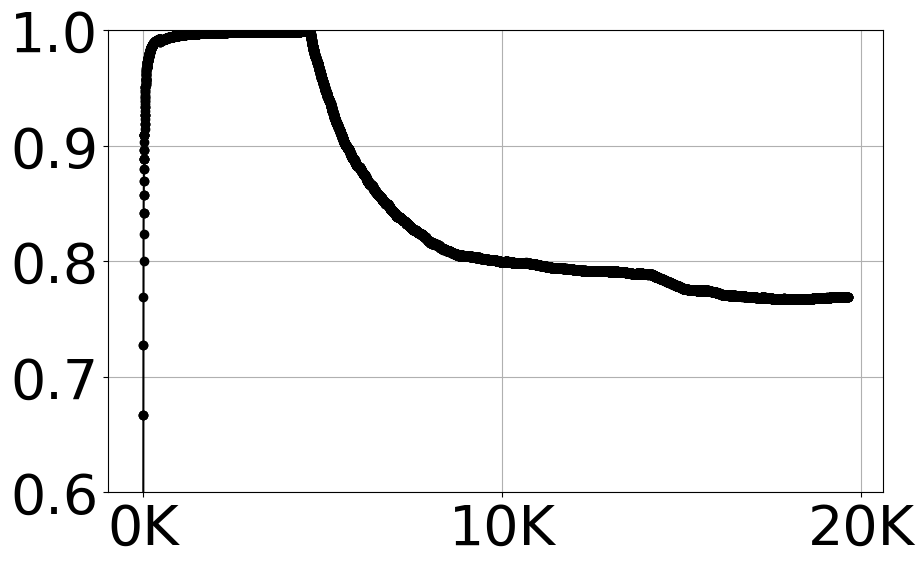}
        \caption{\rev{Worst NB $\rightarrow$ Simulation 2 (mixed dataset).}}
        \label{fig:nb-worst}
        \end{subfigure}
        
        \caption{Comparison between tree-based and non-tree-based streaming ML methods. Top side: best runs. Bottom side: worst runs. X-axis: samples. Y-axis: cumulative \rev{F1-score}. \rev{See \cref{tab:simulations,tab:datasets} for content of simulations.}}
        \label{fig:comparison-tree-nontree}
    \end{minipage}
\end{figure*}

\Cref{tab:simulations} shows how we created the sequence of four different binary classifications in each of the six simulations. The letters E, M and I refer to the source of the traffic of each phase: public datasets Edge-IIoT, MQTTSetand MQTT-IoT respectively (see \cref{subsec:datasets} for details on the datasets). As the table indicates, we alternated between the traffic of different datasets to make the concept drift more intense. We also took care to alternate the type of attack and the balance between benign and malicious traffic, all to maximize data variability and reduce the generalization of previously learned data. \Cref{tab:datasets} shows the details of traffic of all phases in every simulation, including class and type of anomaly, and number of traffic samples.

Since data streams were created with recent public IoT datasets, our proposed simulation structure can serve as a benchmark for other studies evaluating robustness against concept drift.}

In this study, we performed six types of simulations of data streams, each with four phases of different malicious attacks, to cause concept drift in non-adaptive models. \rev{Since data streams possess time dependencies related to the order of the attacks, each dataset was simulated as a data stream in the forward and backward order. To account for variability, each simulation was performed five times (a total of 30 simulations for each ML algorithm). The results in prediction performance were extracted as mean and standard deviation} \revv{\cref{tab:metrics}).}

These precautions in data preparation were necessary because preliminary tests using pure datasets exhibited an artificial generalization of the learned attacks to detect new ones---not because of intrinsic good perfomance of ML algorithms, but because of poor data variability. We concluded that this issue was due to the challenges of creating testbeds complex enough to generate rich datasets. For example, few devices, simple topology and repetitive traffic behavior all cause traffic to be too homogeneous. Therefore, to increase the variability of the simulated data streams, we performed the following:
\begin{enumerate}
\item Data from different datasets were used in the same simulation to diversify traffic properties (e.g., IP addresses, packet sizes).
\item Subsets of different datasets were alternated to maximize drift between phases.
\item Different attack classes were used in the same simulation (DoS, malware, information gathering) to hinder generalization and enhance concept drift.
\item The order of phases of attack were standardized in the three streaming datasets, to improve reproducibility between simulations.
\item Heterogeneous sections of each subset were selected, then benign data was mixed and randomized with malicious data inside each phase to simulate the random order of incoming data streams.
\end{enumerate}

Regarding the simulation of anomaly detection, we implemented a scheme of parallel testing and incremental learning in the three streaming ML models (ARF, Hoeff and NB), as explained in~\cref{subsec:anomalydetector}. We assumed packet labels become available immediately after testing, to contrast continuous model improvement with the less frequent updates of batch models, done through complete retraining. The training phase of the models is summarized below:
\begin{itemize}
\item A batch-trained Random Forest model, validated on phase one, served as a reference, remaining untrained for later phases.
\item Streaming ML models started without pretraining.
\item Each incoming flow sample was first evaluated, then used for supervised incremental training of streaming models.
\end{itemize}

\section{Results}
\label{sec:results}





\begin{table*}[ht]
\caption{\rev{Metrics of detection performance for four ML algorithms over 30 simulations of mixed datasets.}}
\label{tab:metrics}
\centering
\begin{tabular}{lccccccc}
\toprule
\multirow{2}{*}{\textbf{Algorithm}} & \multirow{2}{*}{\textbf{Dataset}}  & \textbf{F1 Score} & \textbf{Accuracy} & \textbf{Precision} & \textbf{Recall} & \textbf{AUC} & \textbf{Bandwidth} \\ 
& & \multicolumn{5}{c}{(Mean $\pm$ Standard Deviation)} & (Mbps) \\ 
\midrule
    \textbf{1. RF} & All & \textbf{0.638 $\pm$ 0.182} & 0.687 $\pm$ 0.247 & 0.659 $\pm$ 0.189 & 0.638 $\pm$ 0.227 & 0.652 $\pm$ 0.182 & 4.112 $\pm$ 0.119\\
    RF best run & MQTT$\rightarrow$IoT & \textbf{0.859} & 1.000 & 0.894 & 0.753 & 0.877 & 3.92 \\
    RF worst run & IoT$\rightarrow$Edge & \textbf{0.380} & 0.352 & 0.365 & 0.413 & 0.368 & 4.1 \\
\midrule
    \textbf{1. ARF} & All & \textbf{0.988 $\pm$ 0.006} & 0.992 $\pm$ 0.004 & 0.989 $\pm$ 0.005 & 0.983 $\pm$ 0.010 & 0.988 $\pm$ 0.005 & 10.857 $\pm$ 0.597 \\
    ARF best run & Edge$\rightarrow$MQTT & \textbf{0.996} & 0.996 & 0.996 & 0.996 & 0.996 & 10.27 \\
    ARF worst run & IoT$\rightarrow$Edge & \textbf{0.978} & 0.978 & 0.979 & 0.977 & 0.979 & 10.49 \\
\midrule
    \textbf{3. Hoeffding T.} & All & \textbf{0.909 $\pm$ 0.074} & 0.920 $\pm$ 0.054 & 0.919 $\pm$ 0.061 & 0.904 $\pm$ 0.109 & 0.918 $\pm$ 0.064 & 17.019 $\pm$ 2.013 \\
    Hoeff best run & Edge$\rightarrow$MQTT & \textbf{0.990} & 0.988 & 0.990 & 0.992 & 0.990 & 12.12 \\
    Hoeff worst run & IoT$\rightarrow$Edge & \textbf{0.703} & 0.865 & 0.764 & 0.592 & 0.755 & 16.1 \\
\midrule
    \textbf{4. Naive Bayes} & All & \textbf{0.832 $\pm$ 0.074} & 0.924 $\pm$ 0.080 & 0.855 $\pm$ 0.068 & 0.765 $\pm$ 0.099 & 0.850 $\pm$ 0.065 & 34.228 $\pm$ 1.003 \\
    NB best run & MQTT$\rightarrow$IoT & \textbf{0.952} & 0.987 & 0.960 & 0.919 & 0.955 & 35.48 \\
    NB worst run & MQTT$\rightarrow$Edge & \textbf{0.769} & 0.758 & 0.768 & 0.781 & 0.768 & 33.71 \\
\bottomrule
\end{tabular}
\end{table*}

\revv{We summarized our results in \cref{fig:comparison-batch-stream,fig:comparison-tree-nontree} and \cref{tab:metrics}. The figures show ML algorithms performing anomaly detection in data streams with concept drift. Fall in performance indicates the occurrence of concept drift and failure to detect new anomalies, while increasing performance indicates adaptation to the new anomalies and successful detection.} 
\Cref{tab:metrics} shows an in-depth evaluation of performance of all algorithms and their computational cost (see ``Algorithms Comparison'' and ``Feasibility'' below).

Before discussing results, it is important to note that streaming anomaly detection \rev{usually} performs worse than batch anomaly detection. Batch algorithms often reach detection scores 0.99, but streaming methods \rev{tend} to not match this because concept drift causes detection failures during the adaptation period. As a result, overall performance \rev{tends} to fall below batch levels.

\subsection{Robustness}
\label{subsec:robustness}

Our evaluation demonstrated the inability of offline batch learning to detect anomalies after concept drift and the robustness of streaming learning algorithms. As mentioned in \cref{subsec:simulation}, when the dataset was not very heterogeneous (e.g. standalone Edge dataset), batch RF still managed to have good detection performance (\cref{fig:RF-edge}), despite the streaming of consecutive malicious traffic that RF was not trained for. ARF performance was similarly very good (\cref{fig:ARF-edge}). However, when datasets were diversified through the measures described in \cref{subsec:simulation}, batch Random Forest (RF) struggled to handle new anomalies and concept drift in general (\cref{fig:RF-edgeMQTT}), while ARF adapted even after initial drops in performance after concept drift, keeping \rev{cumulative F1-score well above 0.90} (\cref{fig:ARF-edgeMQTT}). The results confirmed that batch models struggle with concept drift, and that streaming models can improve performance by means of frequent light-weight updates of the model and detection of concept drift to forget old data.

\subsection{Algorithms comparison}
\label{subsec:algorithm_comparison}

We evaluated the relative strengths of tree-based ML algorithms---the current state-of-the-art of ML-based anomaly detection algorithms--- with non-tree-based methods and confirmed that their strengths apply to IoT traffic streaming anomaly detection. Below we present a discussion of the behavior and performance of every ML algorithm.

\begin{itemize}
    \item \textbf{ARF} showed strong performance with \rev{F1-score of 0.988 ±0.006.} Its detection stability is also shown in its best and worst run (\cref{fig:arf-best,fig:arf-worst}). Such performance is due to the synergy between independently trained trees and independent drift detection for each tree, producing a very efficient incremental adaptation to new data and refined, granular removal of obsolete model parameters. Regarding its computational cost (last column of \cref{tab:metrics}, same for all other algorithms), ARF's bandwidth was \rev{2.5×} smaller than batch RF's, but it may not be viable for online applications. However, in offline applications that process big volumes of data and forbid the complete retraining of models or the storage of data, ARF is ideal to seamlessly update the model with robustness to concept drift.
    \item \textbf{Hoeffding Adaptive Tree} showed relatively good performance with \rev{F1-score of 0.909 ±0.007 and a best run of 0.990 (\cref{fig:hoeff-best}), but displayed stability issues in the standard deviation and a difference of 0.300 between best and worst runs (\cref{fig:hoeff-worst}). It was 4× faster than RF.} The worst run contained imbalanced data (see IoT$\rightarrow$Edge in~\cref{tab:datasets}), which suggests a trade-off between speed and detection robustness. \rev{The biggest limitation was the high standard deviation of recall, a behavior expected from the high variability and concept drift of the mixed datasets (although the average recall itself kept the same value of around 0.900, as the other metrics.}
    \item \textbf{Gaussian Naive Bayes} showed average performance (\cref{fig:nb-best,fig:nb-worst}), with an average \rev{F1-score score of 0.832, standard deviation of ±0.007. Its biggest limitation was the lower recall, deepening the same problem of Hoeffding Tree--less robustness to concept drift. However, as expected from such a simple algorithm, it was the fastest of the methods, with double bandwidth of Hoeffing Tree, triple of ARF and nine times of RF. This trade-off makes it a desirable faster algorithm, in case the lower but still competitive F1-score of 0.830 is acceptable.}
\end{itemize}

In summary, tree-based algorithms confirmed their better robustness in comparison with the analyzed non-tree-based ones, with specific application possibilities for all tested algorithms.

\subsection{Feasibility}
\label{subsec:feasibility}

In this study we aimed at replicating the requirements of a deployment scenario by processing traffic in raw pcap files and calculating the total computational cost of the pipeline as the bandwidth of traffic. \rev{Bandwidth was calculated by simply counting the amount of megabits in the dataset and dividing it by the amount of time taken to process it through the whole pipeline.} in The last column of \cref{tab:metrics} shows the output bandwidth of each ML method used inside our anomaly detection framework. While current bandwidth (\rev{10-34 Mbps}) is insufficient for deployment, fast methods like Hoeffding Adaptive Tree show potential for real-time anomaly detection.

\subsection{Limitations}
\label{subsec:limitations}

Our assumption regarding the immediate availability of labels for incremental training is idealistic. It was made to display the biggest advantage of streaming learning compared to batch learning: the continuous training of the model sample by sample. However, real scenarios will present a bottleneck in the frequency with which security operators can label post-tested data and give it to the model for incremental training. Therefore, our results are best described as the upper bounds in performance, while the observed performance of batch learning is the lower bound, since it represents the performance of a model when it is never updated. Another limitation of the study is the maximum output bandwidth of our anomaly detection framework: \rev{34 Mbps}. To achieve high-throughput, our Python-based implementation will require further optimization or porting to a faster language.





\section{Conclusion}
\label{sec:conclusion}



In this paper, we evaluated streaming IoT traffic anomaly detection and demonstrated the importance of streaming ML algorithms in adapting to concept drift. We also examined data requirements for realistic simulations and found that current datasets lack sufficient heterogeneity. To address this, we mixed three types of datasets in pairs and created enough heterogeneity for our evaluations.

Our results confirmed the loss of robustness in batch Random Forest against concept drift. Regarding the comparison of streaming ML algorithms, Adaptive Random Forest was very robust \rev{(99\% performance) and stable. It was two times faster than Random Forest, but showed the highest computational cost between the streaming algorithms. Hoeffding Adaptive Tree provided a fast tree-based alternative (four times faster than RF) with strong concept drift adaptation, but had slightly less stable recall for heterogeneous malicious traffic. Gaussian Naive Bayes showed average results, but potential for faster detections where 80\% of detection performance is acceptable.} In general, tree-based models outperformed non-tree-based ones.



\section*{Acknowledgment}
We thank Uyen Do, Manuel Poisson and Jaroslav Pesek for their feedback during meetings. The work is partially supported by JST CREST JPMJCR21M3.

\vspace{2mm}

\noindent\textbf{AI usage disclosure:} The original draft was written without any AI assistance. ChatGPT was used to review grammar and simplify sentences in all paper sections.

\noindent\textbf{Software availability:} 
\siuru's expanded source code is publicly available on \url{https://github.com/rodrigo-carnier/siuru}.


\bibliographystyle{IEEEtran}
\bibliography{IEEEabrv,10-ref-general,11-ref-solutions,15-ref-old}


\end{document}